\documentclass[conference]{IEEEtran}
\IEEEoverridecommandlockouts
\usepackage{amsmath,amssymb,amsfonts}
\usepackage{algorithmic}
\usepackage{graphicx}
\usepackage{textcomp}
\usepackage{xcolor}
\usepackage[nottoc]{tocbibind}
\usepackage[round]{natbib}
\usepackage{romannum}
\usepackage{multirow}
\usepackage{adjustbox}
\usepackage{algorithm}

\def\BibTeX{{\rm B\kern-.05em{\sc i\kern-.025em b}\kern-.08em
    T\kern-.1667em\lower.7ex\hbox{E}\kern-.125emX}}
\begin{document}

\title{Online Hand Gesture Recognition Using 3D Convolutional Neural Networks\\
\thanks{https://github.com/KingQino/Online-Hand-Gesture-Recognition}
}

\author{\IEEEauthorblockN{Yinghao Qin, Tijana Timotijevic$^\star$}
\IEEEauthorblockA{\textit{School of Electronic Engineering and Computer Science} \\
\textit{Queen Mary, University of London}\\
London, UK \\
\{y.qin, t.timotijevic\}@qmul.ac.uk}
\thanks{$^\star$Corresponding author}
}

\maketitle

\begin{abstract}
In human computer interaction, real-time detection and classification of dynamic hand gestures is challenging as: 1) the system must run in a real-time video stream and there is no noticeable lag in response after performing a gesture; 2) there is a large difference in how people perform gestures, making recognition more difficult. In this paper, an online hand gesture recognition system is proposed, which is able to localize gestures in real-time video stream and recognize what these gestures are. To improve the robustness of the system, the sliding window approach is used to refine results from multiple windows. All of the models in my project are trained on Jester database, achieving 98+\% accuracy for detector and 90+\% accuracy for classifier. For the overall performance of the system, the best group can respond within three seconds and reach 37.5\% Levenshtein accuracy on the homemade dataset. The project codes used in this work are publicly available.
\end{abstract}


\section{Introduction}
Recently, hand gesture recognition has attracted more and more attention in human machine interaction. It can be applied in numerous fields, such as sign language recognition, in-vehicle gestural interface, video surveillance system, gaming and virtual reality control etc. By and large, there are two kinds of hand gesture recognition technologies. First, wearable sensor devices \citep{1, 2, 3} can precisely track information such as hand position, movement velocity, acceleration etc. From the data, hand gesture can be inferred easily, whereas the approach is unrealistic for most people due to its inconvenience and additional cost. Second, computer-vision based methods require large amount of data to train the system for generalizing the unknown scenarios instead of requiring extra devices. Nowadays, the most popular approach in the field of dynamic hand gesture recognition is using deep convolutional neural networks (CNN) to learn and generalize visual information e.g., \citep{4,5,6,7,8}.

Although there have been a lot of deep learning researches about dynamic gesture recognition, most of them just focus on how to learn more representative spatiotemporal features in pre-segmented video clips. Indeed, this has spawned a number of works with high classification accuracy on many well-known databases. However, few researchers put their sights on the real-world application of hand gesture recognition. This paper is an attempt to bridge the gap between them. 

Notably, dynamic hand gesture recognition in the real world still faces numerous challenges. Firstly, how to identify when a gesture begins and ends in the video, which is actually a temporal action localization/recognition task. In addition, the system should be robust to uncertain environment such as illumination, background, occlusion. Furthermore, when people perform hand gestures, the temporal scale is required to be considered since people gesture with different duration. Finally, we should take cost and efficiency into account as well, i.e. the system should cost relatively little computational resources and react quickly. 

There are basically two core tasks in real-world system for dynamic hand gesture recognition - temporal gesture localization and gesture classification. The former is designed to localize the gesture proposals in the video stream, and the latter aims to gain a high classification accuracy with low consumption of resources.

Overall, the contributions of this project are mainly in three aspects. 1) Inspired by \citet{4}, an online hand gesture recognition system is designed, which can respond to hand gestures within 3 seconds and achieve 37.5\% Levenshtein accuracy in self-collected dataset; 2) the framework is systematically evaluated from components to the entire system; 3) a couple of 3-dimensional (3D) CNN models trained on Jester database \citep{9} are obtained.

The rest of the paper is organized as follow. Section \Romannum{2} reveals the related work regarding action and gesture recognition. In section \Romannum{3}, the online hand gesture recognition system is described in detail. Next, section \Romannum{4} presents the experiment results and evaluations. The paper is then concluded in section \Romannum{5}. Lastly, section \Romannum{6} gives some ideas about future work.

\section{Related work}

Many handcrafted spatiotemporal features for visual data analysis are first developed in the area of action and gesture recognition. These methods typically extract features such as shape, appearance, optical flow to perform gesture classification. In their paper, \citet{10} state an exemplar-based approach for dynamic hand gesture recognition, which first detects salient regions from motion divergence fields and then extracts local descriptors to index gesture from the pre-trained vocabulary. \citet{11} systematically evaluate the efficacy of action recognition using 7 low-level static and dynamic features, all of which are represented by BoW descriptors and an SVM approach is used to classify gesture. \citet{12} apply a descriptor combining RGB and depth features to classify hand gestures, and meanwhile propose an architecture with two independent modules to detect hand gestures and recognize them. \citet{13} introduce a video representation by deriving feature points matches from multiple frames using SURF descriptors and dense optical flow, and also compare two feature encoding approach - Fisher vector and Bag-of-words histogram.

After that, due to the breakthrough of CNNs in static images, many people try to apply 2D CNNs in video analysis. The typical method is to first extract features from individual frame using 2D CNNs, then take fusion measures to track motion of objects in time dimension, e.g., \citep{14,15}. In addition, \citet{15} also apply CNNs in stacked optical flow maps from consecutive frames to generalize action in video. Long term recurrent convolutional networks are proposed by \citet{16}, where 2D CNNs are used to derive visual features from each frame and these features are fed into a stack of recurrent sequence models (LSTMs) to generalize actions. Temporal segment network (TSN) is proposed by \citet{17}, based on the idea of long-range temporal structure modeling, uses 2D CNNs to extract spatial information from color modality and temporal information from optical flow modality in uniformly distributed samples.

Gradually, some researchers begin to explore whether they could use a single CNN framework to learn spatiotemporal features instead of using multiple 2D CNNs. To the best of my knowledge, the idea of 3D CNNs for action recognition is first stated by \citet{18}. \citet{19} propose a homogeneous convolutional 3D architecture with small 3$\times$3$\times3$ convolution kernels, where the input data has 4 dimensions including the number of channels, the number of frames in a clip, the height and width of the frame. \citet{20} fuse multi-modalities data to train the 3D CNN model for hand gesture recognition of driver. In the following, 3D ResNets, ResNeXts, DenseNets are subsequently used for spatiotemporal information learning \citep{21,22}. 

\begin{figure}[htbp]
\centerline{\includegraphics[scale=0.6]{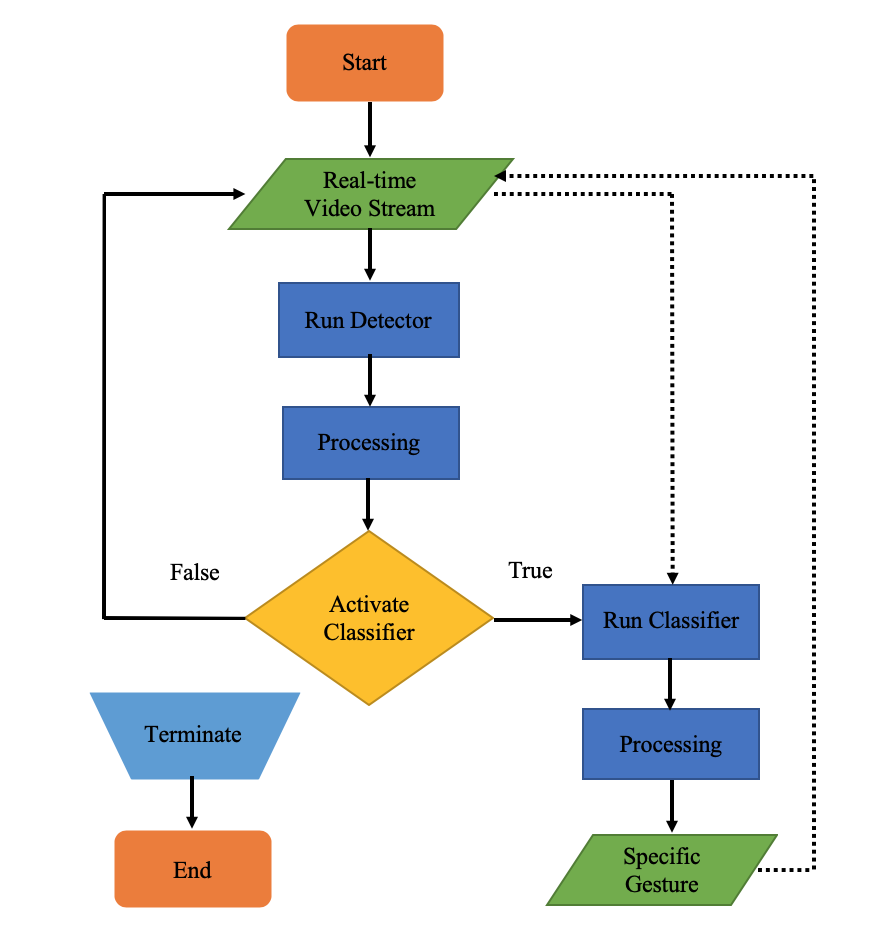}}
\caption{The flow chart of the online hand gesture recognition system.}
\label{fig1}
\end{figure}

In addition, there are other developments in hand gesture and action recognition. Firstly, many hand gesture and action databases have appeared in recent years, such as Sports-1M \citep{14}, NVGesture \citep{5}, EgoGesture \citep{zhang2018egogesture}, Jester \citep{9} etc. \citet{8} propose to fuse motion information into still images as the spatiotemporal representatives of an action, namely Motion Fused Frames (MFFs). They claim that it is the first time that data-level fusion has been applied to deep-learning based motion and gesture recognition. \citet{5} state a system, simultaneously detecting and classifying hand gesture from online video stream, which is composed of a recurrent 3D CNN and a connectionist temporal classifier, CTC. A real-time hand gesture recognition system is proposed by \citet{4}, which contains a detector and a classifier. In addition, they also recommend the Levenshtein distance as the evaluation metric.

\section{Approach}
In this section, I elaborate on the hierarchical architecture which is used to recognize dynamic hand gesture, and then describe the detection and classification methods. By the way, the Levenshtein evaluation metric is also introduced.

\subsection{Architecture}\label{AA}
The subsection first demonstrates the overall architecture and sliding window approach of the system and then analyzes qualities the system should meet. Following that, the evaluation metric on the overall system is introduced.

\begin{table}[htbp]
\caption{The Architectures of ResNet-10 and ResNeXt-101}
\begin{center}
\begin{adjustbox}{angle=90}
\begin{tabular}{c|c|c|c}
\hline
layer & output & ResNet-10 & ResNeXt-101 \\ \hline
conv1 & d×56×56 & 3×7×7, 16, stride=(1,2,2) & 3×7×7, 64, stride=(1,2,2) \\ \hline
\multirow{5}{*}{conv2} & \multirow{5}{*}{d/2×28×28} & 3×3×3 max pool, stride=2 & 3×3×3 max pool, stride=2 \\ \cline{3-4} 
 &  & \multirow{4}{*}{$\left[{3\times3\times3,\ 16\atop3\times3\times3,\ 16}\right]\times1,\ 16$} & \multirow{4}{*}{$\left[\begin{matrix}1\times1\times1,\ 128\\3\times3\times3,\ 128,\ C=32\\1\times1\times1,\ 128\\\end{matrix}\right]\times3,\ 128$} \\
 &  &  &  \\
 &  &  &  \\
 &  &  &  \\ \hline
\multirow{4}{*}{conv3} & \multirow{4}{*}{d/4×14×14} & \multirow{4}{*}{$\left[{3\times3\times3,\ 32\atop3\times3\times3,\ 32}\right]\times1,\ 32$} & \multirow{4}{*}{$\left[\begin{matrix}1\times1\times1,\ 256\\3\times3\times3,\ 256,\ C=32\\1\times1\times1,\ 512\\\end{matrix}\right]\times3,\ 256$} \\
 &  &  &  \\
 &  &  &  \\
 &  &  &  \\ \hline
\multirow{4}{*}{conv4} & \multirow{4}{*}{d/8×7×7} & \multirow{4}{*}{$\left[{3\times3\times3,\ 64\atop3\times3\times3,\ 64}\right]\times1,\ 64$} & \multirow{4}{*}{$\left[\begin{matrix}1\times1\times1,\ 512\\3\times3\times3,\ 512,\ C=32\\1\times1\times1,\ 1024\\\end{matrix}\right]\times3,\ 512$} \\
 &  &  &  \\
 &  &  &  \\
 &  &  &  \\ \hline
\multirow{4}{*}{conv5} & \multirow{4}{*}{$\lceil${d/16}$\rceil$×4×4} & \multirow{4}{*}{$\left[{3\times3\times3,\ 128\atop3\times3\times3,\ 128}\right]\times1,\ 128$} & \multirow{4}{*}{$\left[\begin{matrix}1\times1\times1,\ 1024\\3\times3\times3,1024,\ C=32\\1\times1\times1,\ 2048\\\end{matrix}\right]\times3,\ 1024$} \\
 &  &  &  \\
 &  &  &  \\
 &  &  &  \\ \hline
\multirow{2}{*}{} & \multirow{2}{*}{1×1×1} & \multirow{2}{*}{average pool, 128-d fc, softmax} & \multirow{2}{*}{average pool, 2048-d fc, softmax} \\
 &  &  &  \\ \hline
\multicolumn{2}{c|}{\# params} &  {0.90×10$^{\mathrm{6}}$}&  {70.49×10$^{\mathrm{6}}$}\\ \hline
\end{tabular}
\end{adjustbox}
\label{tab1}
\end{center}
\end{table}

The workflow of the hand gesture recognition system is designed as shown in Fig.~\ref{fig1}. It consists of two phases - detection and classification. In detection stage, real-time video stream captured by RGB sensors is fed into the system, meanwhile the detector runs to distinguish between ‘gesture’ and ‘no gesture’ classes from the video stream, and then the detection results are further processed to improve the detection performance. Afterwards, we activate the classifier if ‘gesture’ exists based on the detection results, otherwise update the video stream for the next round of detection. When the system enters the classification phase, video stream is fed into the classifier to generate probabilities of different gesture classes, which are then further processed to predict predefined gestures. In addition, the system can be manually terminated at any time the system is running. 

\begin{figure}[htbp]
\centerline{\includegraphics[scale=0.8]{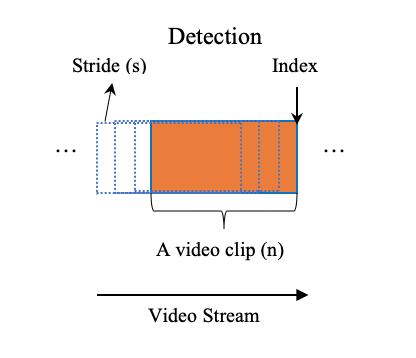}}
\caption{Process video stream using sliding window in detection stage.}
\label{fig2}
\end{figure}

\begin{figure}[htbp]
\centerline{\includegraphics[scale=0.8]{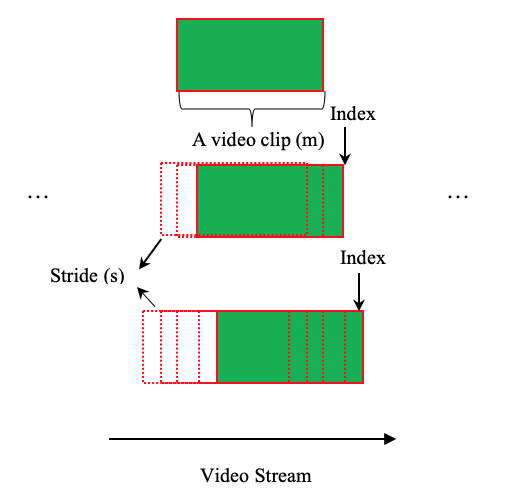}}
\caption{Process video stream using sliding window in classification stage.}
\label{fig3}
\end{figure}

The sliding window method is applied to further process the original results from detector and classifier. Intuitively, using data from multiple sliding windows can increase prediction confidence as these windows cover more useful information. For the input data (real-time video stream), consecutive frames are encapsulated into a clip as a 4-dimensional input value. For the sake of simplicity, we represent video clips with the sign of $c\times d\times h\times w$, where $c$ refers to the number of channels, $d$ is the depth of the clip, and $h$ and $w$ represents the height and width of the frame respectively. As shown in Fig.~\ref{fig2} and Fig.~\ref{fig3}, the detector and classifier are fed by video clips with size $n$ and $m$ respectively. In details, a detector clip contains only 8 frames, whereas a 16-frame or 32-frame video clip is used in classification stage. For each round of detection, four successive clips with stride of s are used to judge whether there is a gesture in the area, as shown in Fig.~\ref{fig2}. In this experiment, the stride s of detector is set as 1. 
In the classification stage, multiple windows are used to decide which specific gesture it is. This specific method is explained in the classification subsection.

Besides, the dynamic hand gesture recognition system should meet certain qualities. First, it should be highly efficient, i.e. the system can respond as quickly as possible. When a gesture ends, the system can return recognition result in no more than three seconds, and we define it is highly efficient. Second, the system should have an acceptable accuracy in recognition, namely 30\% Levenshtein accuracy. Third, the system should be compatible with different detectors and classifiers, which allows different models to be changed easily. 

Levenshtein accuracy is proposed by \citet{4} for online evaluation, which can measure the difference between the prediction results and the ground truth. Levenshtein accuracy of the recognition is calculated by the corresponding Levenshtein distance and the size of the ground truth samples. Levenshtein distance is originally a string metric for measuring the difference between two sequences \citep{27}. In details, it is the minimum number of single-character edits (i.e. intersections, deletions or substitutions) required to change one sequence into the other. Specifically, the metric takes into account missing detections, misclassifications and multiple recognitions in the field of gesture recognition. The Levenshtein accuracy calculation formula is presented in \eqref{eq1}.

\begin{equation} 
LA(prediction) = (1- \frac{LD(target,~prediction)}{length(target)}) \times 100\%
\label{eq1}
\end{equation}

Where $LA$ denotes the Levenshtein accuracy of the prediction, $LD$ represents the Levenshtein distance between the true target sequence and the prediction sequence, $length(target)$ refers to the length of the true target sequence.

\subsection{Detection}

The goal of detection phase is to distinguish gesture and no gesture from the real-time video stream, which is a preparation for the next classification stage. Since the performance of the detector is closely related to the performance of the whole system, thus the detector is required to achieve a high accuracy and be as lightweight as possible. Two steps are taken to attain the goal. Firstly, we need to train a detector, which is actually a binary classification model. Next, after getting this raw detection results, we need to do some post-processing to improve the stability of the results.

\begin{figure}[htbp]
\centerline{\includegraphics[scale=0.6]{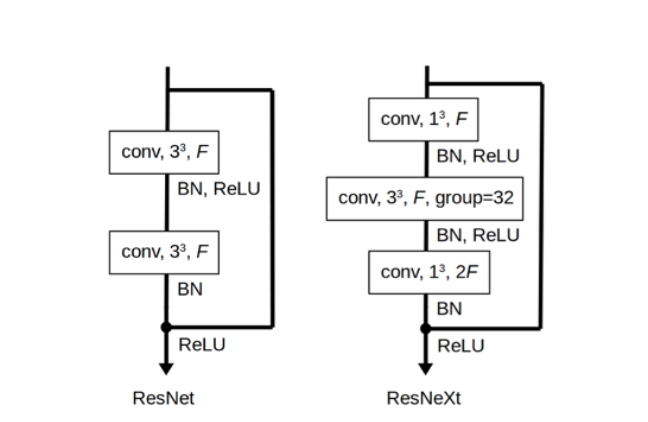}}
\caption{The basic blocks for ResNet-10 and ResNeXt-101. $conv,\ 3^3$ or $1^3$ represents convolution operation using $3\times3\times3$ or $1\times1\times1$ kernel. F, BN, ReLU and group denotes the number of channels, batch normalization, rectified linear unit and cardinality respectively \citep[p. 4]{4}.}
\label{fig4}
\end{figure}

A 10-layer ResNet is chosen as the detector owing to its low complexity and simplicity. In details, the 10-layer ResNet model only has less than 1 million parameters, as shown in Table~\ref{tab1}. Indeed, \citet{21} have tried to apply 3D residual networks for action recognition and got exciting results. The basic block of ResNet is shown in the left part of Fig.~\ref{fig4}, where the input clip flows into the stacked layers and a shortcut connection, and their outputs finally are combined to generate the new output. The detailed architecture of ResNet-10 in our case is demonstrated in Table~\ref{tab1}. A $3\times8\times112\times112$ clip inputs the system to generate a 128-dimensional vector. Notably, there is a ceiling brackets in ‘conv5’ layer for the output of ‘d/16’, ensuring it is integer value. 

After the detector gives multiple detection results using sliding window method, two different methods are employed to make the final detection decisions - the first one utilizes the raw probabilities of the detector predictions and the other one just uses the specific prediction results. Due to the sliding window approach, there is a queue containing $k$ predictions (probabilities or specific prediction results). The size of the queue is selected as 4, as shown in Fig.~\ref{fig2}. For the first method, we calculate the average of the values in the queue, and then select the maximum value as the prediction results. For the second one, we just judge whether all of the values are ‘gesture’ class. If so, the system activates the classifier, otherwise continue the next round of detection.

\begin{table}[htbp]
\caption{The Architectures of C3D}
\begin{center}
\scalebox{0.95}{
\begin{tabular}{c|c|c}
\hline
layer & output & C3D \\ \hline
conv1a & d×112×112 & 3×3×3, 64, stride=1 \\ \hline
\multirow{2}{*}{conv2a} & \multirow{2}{*}{d×56×56} & 1×2×2 max pool, stride=(1,2,2) \\ \cline{3-3} 
 &  & 3×3×3, 128, stride=1 \\ \hline
\multirow{3}{*}{conv3a,b} & \multirow{3}{*}{d/2×28×28} & 2×2×2 max pool, stride=2 \\ \cline{3-3} 
 &  & \multirow{2}{*}{{[}3×3×3, 256, stride=1{]}×2} \\
 &  &  \\ \hline
\multirow{3}{*}{conv4a,b} & \multirow{3}{*}{d/4×14×14} & 2×2×2 max pool, stride=2 \\ \cline{3-3} 
 &  & \multirow{2}{*}{{[}3×3×3, 512, stride=1{]}×2} \\
 &  &  \\ \hline
\multirow{3}{*}{conv5a,b} & \multirow{3}{*}{d/8×7×7} & 2×2×2 max pool, stride=2 \\ \cline{3-3} 
 &  & \multirow{2}{*}{{[}3×3×3, 512, stride=1{]}×2} \\
 &  &  \\ \hline
\multirow{2}{*}{} & \multirow{2}{*}{d/16×4×4} & 2×2×2 max pool, stride=2 \\ \cline{3-3} 
 &  & max pool, 4096-d fc6\&7, dropout=0.5, softmax \\ \hline
\multicolumn{2}{c|}{\# params} & 78.11×10$^{\mathrm{6}}$ (16-frame), 111.67×10$^{\mathrm{6}}$ (32-frame) \\ \hline
\end{tabular}}
\label{tab2}
\end{center}
\end{table}

In addition, in offline experiment, some measures are also taken to refine some proposal areas with gesture. After using the sliding window approach to get multiple original proposals, they will be merged and amended. In details, two consecutive areas are merged into a larger region when the intersection between the two proposal areas is less than the threshold (which is set as 4 in the experiment), and meanwhile some amendments are made to the new proposals. Finally, these proposal areas are used for classification task.

\subsection{Classification}
Similar to the detection phase, two steps are taken to get the final classification decision – classifier run and post-processing. For the former, different classifiers are trained on Jester dataset with 16-frame clip and 32-frame clip. For the latter, we take advantage of the raw classification probabilities to improve the performance.

For classifier, C3D and ResNeXt-101 models are selected as our classifier models. \citet{19} propose C3D model, which is one of the initial 3D CNNs and performs well in previous work. The ResNeXt-101 model is first proposed by \citet{24}, by increasing cardinality to gain higher accuracy instead of going deeper or wider. In other words, the ResNeXt structure can improve the accuracy without increasing the complexity of parameters, and meanwhile reduce the number of super parameters. \citet{22} state that 3D ResNeXt to learn spatiotemporal features for action recognition. The architectures of the two model are demonstrated in Table~\ref{tab1} and~\ref{tab2}.

\begin{figure}[htbp]
\centerline{\includegraphics[scale=0.23]{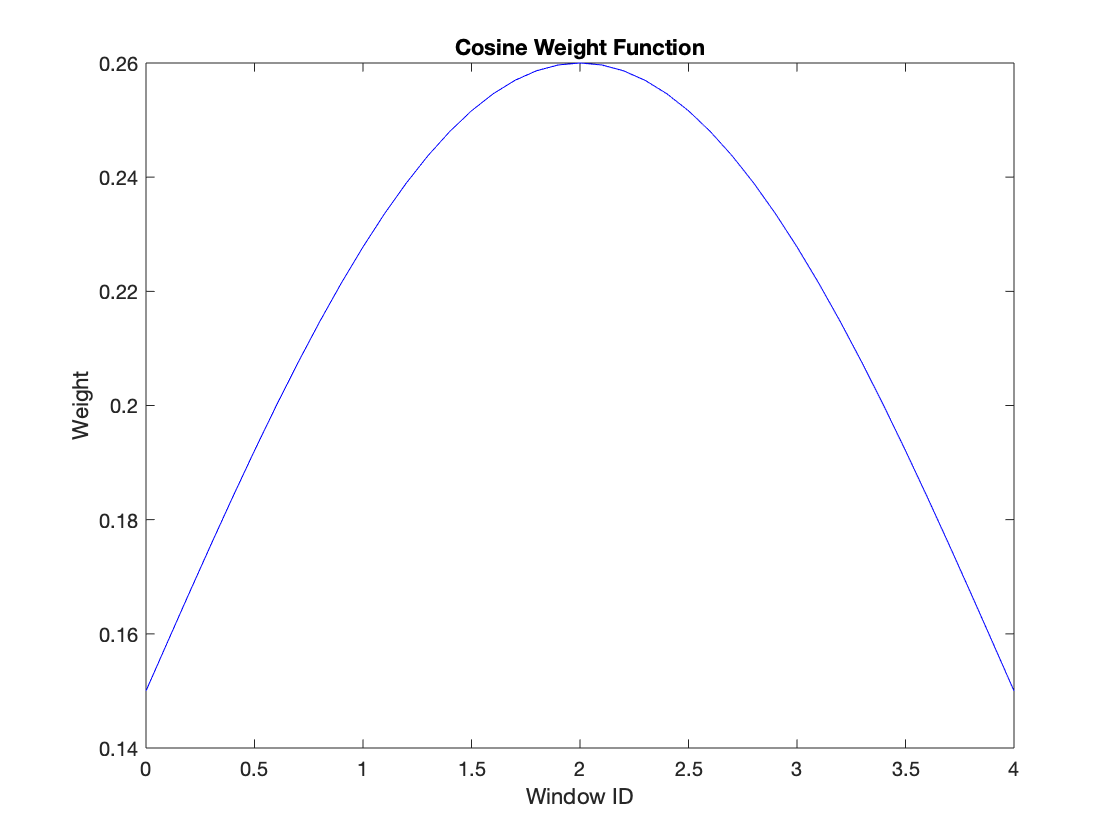}}
\caption{The cosine weight function according to \eqref{eq2}.}
\label{fig5}
\end{figure}

In order to find the balance between the classification accuracy and the reaction time, three different kinds of sliding windows are designed. As shown in Fig.~\ref{fig3}, 1, 3, 5 windows of video clips are used to make the final classification decision. The stride of these windows is set as 1. Notably, the windows in the middle should gain more confidence, i.e. the weights corresponding to the middle window should be larger. This is because, according to the theory \citep{25,26}, there are three temporally overlapping phases in dynamic hand gestures: preparation, nucleus, and retraction, of which the nucleus is the most discriminative. The sliding windows we use should try the best to cover the nucleus part of hand gestures, which probably result in a good performance. Thus, for the 3 windows method (the middle part in Fig.~\ref{fig3}) the weights are set as $0.3, 0.4, 0.3$ respectively. And for the 5 windows method (the bottom part in Fig.~\ref{fig3}), the weight is formulated as a cosine function as shown in \eqref{eq2}.

\begin{equation} 
w_i=0.11\times cos\left(\frac{\pi}{4}\cdot x-\frac{\pi}{2}\right)+0.15
\label{eq2}
\end{equation}

Where $w_i$ is the weight of the $i^{th}$ window and $x$ is the window identifier. And the weight function plot is shown in Fig.~\ref{fig5}. 

Also, the idea of early detection and late detection is used in the project as well. The early detection is when the system responds to a gesture before it is actually over, i.e. zero or negative lag detection. The late detection is a compensation for assuring each gesture should be detected if the early detection does not be activated, which ensures a very high recall rate. Generally, the early detection is more confident than the late detection.

The pseudocode of classification strategy is described in Algorithm~\ref{alg1}. For each window, the raw probabilities produced by the classifier are multiplied by the corresponding weight, and then the results are accumulated. From the generating list, the two highest values are selected. When the difference between the two values is greater than the threshold $\tau_{early}$, then the early detection is activated; if the early detection is not triggered and the maximum value is greater than the threshold $\tau_{late}$, the late detection is triggered. The $\tau_{early}$ and $\tau_{late}$ are set as 0.6, 0.2 respectively in the experiment.

\begin{algorithm}[htbp]
\caption{Classification}
\begin{algorithmic}[1]
\REQUIRE Incoming video frames.
\ENSURE The specific hand gesture class.
\STATE Initialize $weigthprobs$ with [0, ... ,0]
\FOR {each window $w_i$}
\STATE $probs_i \leftarrow classifier(clip_i)$
\STATE $\alpha_i \leftarrow weight_i  \times probs_i$
\STATE $weigthprobs=weigthprobs+\alpha_i $
\ENDFOR
\STATE $(max_1, max_2) = max [weigthprobs]_2$
\IF {$(max_1 - max_2) \geq \tau_{early}$}
\STATE $state \leftarrow "early \  detection"$
\RETURN gesture with $max_1$
\ELSIF{$max_1\geq \tau_{late}$}
\STATE $state \leftarrow "late \  detection"$
\RETURN gesture with $max_1$
\ENDIF 
\end{algorithmic}
\label{alg1}
\end{algorithm}

\section{Experiment}

All the models are trained from sratch on Quadro RTX 6000 processor, with 100 training epochs. The code is programmed in Python and Pytorch. The detailed setting for these models is described as follows. Cross entropy loss function and stochastic gradient descent optimizer are taken for all the models. For C3D model, the initial gradient learning rate is set as ${10}^{-3}$, the momentum is $0.9$ and the weight decay is $5\times{10}^{-4}$. In addition, the learning rate will divide by 10 per 10 epochs. For ResNeXt-101 model, I set the initial learning rate is ${10}^{-2}$, the momentum is $0.9$ and the weight decay is ${1\times10}^{-3}$. Besides, the learning rate will divide by 10 per 15 epochs. 

\subsection{Dataset}

The Jester database \citep{9} of hand gestures, has an about 150,000 examples. They are proposed to split into train, validation, test set using the ratio 8:1:1. In the dataset, there are totally 27 classes, each of which has around 4,000 video clips, except for the class ‘Doing other things’ with nearly 9,500 clips. Particularly, the ‘Doing other things’ class is the contrast category with the collection of various actions, such as stretching, yawing and playing with hair. For each video clip, its average duration is 3 seconds i.e. roughly 36 frames. 

\begin{figure}[htbp]
\centerline{\includegraphics[scale=0.5]{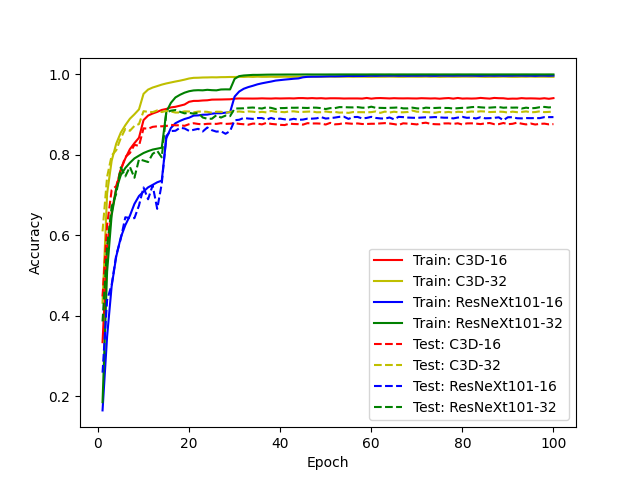}}
\caption{The accuracy comparison of classifiers during the training.}
\label{fig6}
\end{figure}

\begin{figure}[htbp]
\centerline{\includegraphics[scale=0.5]{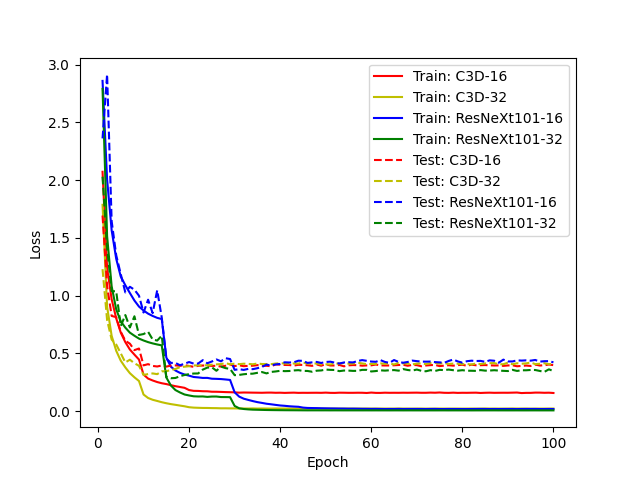}}
\caption{The loss comparison of classifiers during the training.}
\label{fig7}
\end{figure}

\subsection{Detector}
In order to train the detector, a subset is collected from the Jester database and split into ‘Gesture’ and ‘No gesture’ two classes. In details, all of the data in the ‘No gesture’ class in Jester dataset is first collcted as the new ‘No gesture’, and then randomly pick the same number of data from all of the other classes as the ‘Gesture’ class. In the end, there are 8,556 train data and 1,066 test data in the subset, in the ratio of about 8:1, which is exactly proposed by the official Jester dataset paper. Finally, the detector reaches a highest test accuracy of 98.22\% on the subset.

\subsection{Classifier}

According to Fig.~\ref{fig6}, C3D-16 performs the worst due to the lowest test accuracy after these models reach stable. Also, it is notable that the train accuracy difference between C3D-16 and C3D-32 is big, but that between ResNeXt101-16 and ResNeXt101-32 is almost zero after 30 epochs. It seems that training with a larger size of clip does not improve the performance of ResNeXt101, but the test accuracy difference is obvious between ResNeXt101-16 and ResNeXt101-32. Finally, the performances of all the models stabilize from around ${40}^{th}$ training epoch to the training end i.e. there is no overfitting, demonstrating that the penalty factor of the optimizer I set is reasonable.

\begin{table}[htbp]
\caption{The highest test accuracy of models}
\begin{center}
\begin{tabular}{ccc}
\hline
Model & Input (clip) & The Highest Accuracy \\ \hline
\multirow{2}{*}{C3D} & 16-frame & 88.00\% \\
 & 32-frame & 91.03\% \\
\multirow{2}{*}{ResNeXt101} & 16-frame & 89.47\% \\
 & 32-frame & 91.97\% \\ \hline
\end{tabular}
\label{tab3}
\end{center}
\end{table}

\begin{table}[htbp]
\caption{Test accuracy comparison on Jester database}
\begin{center}
\begin{tabular}{lc}
\hline
Method & Accuracy \\ \hline
20BN’s Jester System \citep{9} & 82.34\% \\
C3D & 91.03\% \\
ResNeXt101 & 91.97\% \\
TRN \citep{28} & 94.78\% \\
SSNet RGB resnet \citep{29} & 95.79\% \\
Motion Fused Frames \citep{8} & 96.28\% \\
Deformable ResNeXt101\citep{30} & 96.60\% \\ \hline
\end{tabular}
\label{tab4}
\end{center}
\end{table} 

Moving onto Fig.~\ref{fig7}, C3D generalize faster than ResNeXt101, since C3D reach stable only using nearly 20 epochs whereas ResNeXt101 spending about 30 epochs. This is probably because C3D has far more parameters than ResNeXt101, for example C3D-32 has over 0.5 times more parameters than ResNeXt101-32. Moreover, there are fluctuations in ResNeXt101 loss curve between ${10}^{th}$ and ${20}^{th}$ epoch, which is probably because for the initial learning rate (${10}^{-2}$) of ResNeXt101 is higher than that of C3D (${10}^{-3}$) and the decay step of the learning rate (15) is larger than that of C3D (10).

According to Table~\ref{tab3}, ResNeXt101 architecture with 32-frame input gains the highest test accuracy (91.97\%) among these models. Although the accuracies of C3D-32 and ResNeXt101-32 are extremely close, ResNeXt101-32 has only 70.49 M parameters which is roughly a third less than the parameter numbers of C3D-32.

Table~\ref{tab4} compares my classifier with other classification methods on Jester database. The accuracies of both my models are around 9\% higher than that of the baseline model - 20BN's Jester System. Although my method does not perform better than others in terms of accuracy on Jester database, that is not the concern of my project. The focus of my project is on the real-world application. By the way, it is worth mentioning that some of their methods are too complex and unrealistic to be applied in real world, e.g. \citep{8}.

Overall, ResNeXt101 with a 32-frame input clip can achieve the best performance among these groups, having the highest test accuracy and the smallest parameters. 

\subsection{Entire evaluation}

\begin{table}[htbp]
\caption{The average duration for each stage}
\begin{center}
\begin{tabular}{ccc}
\hline
Detection & \multicolumn{2}{c}{Classification} \\ \hline
\multirow{3}{*}{0.91 s} & 1 window & 1.16 s \\
 & 3 windows & 3.26 s \\
 & 5 windows & 5.33 s \\ \hline
\end{tabular}
\label{tab5}
\end{center}
\end{table}

\begin{table}[htbp]
\caption{The Levenshtein accuracy comparison.}
\begin{center}
\begin{tabular}{ccccc}
\hline
          & C3D-16  & C3D-32  & ResNeXt101-16 & ResNeXt101-32 \\ \hline
1 window  & 27.50\% & 25\%    & 37.50\%       & 25\%          \\
3 windows & 28.75\% & 18.75\% & 35\%          & 25\%          \\
5 windows & 23.75\% & 18.75\% & 33.75\%       & 17.50\%       \\ \hline
\end{tabular}
\label{tab6}
\end{center}
\end{table}

For using the Levenshtein metric to evaluate the entire system, a small dataset is collected. Limited to conditions, the dataset only contains 20 videos, each of which has 4 gestures performed in 20 seconds. In other words, there are totally 80 gestures in 400 seconds video time used for evaluation. By the way, the entire evaluation experiment is made in Quad-Cord Intel Core i7 processor.

Table~\ref{tab5} reveals the average duration per round of detection and classification. It is clear that each round of detection costs no more than 1 second, whereas each round of classification lasts at least 1 second. Besides, the classification lasts longer with more sliding windows. These two phases are main components of the total response time of the system. Thus, when a gesture is performed, the system will cost at least 2 seconds to give the recognition results. However, it actually takes about 3 seconds to recognize that when using the 1-window classification method, because data capture, redundancy processing and so on take time as well. 

From Table~\ref{tab6}, astonishingly, both C3D and ResNeXt101 with 16-frame input generally performs better than those with 32-frame input in terms of Levenshtein accuracy. Although the classifiers with 32-frame input have higher classification accuracy, they make less contribution to the overall performance of the system. This is probably because larger size of input covers more distractions though it has a bigger probability of covering more useful information. In addition, using more classification windows seems to harm the overall performance of the system, which is probably because more windows not only enhance the exact gesture information but also strengthen the distractions. To sum up, using larger size of input data and more windows does not benefit the overall performance of the system, which is inconsistent with our intuitiveness.

Overall, the online system, consisting of a ResNet10 detector with 8 frame input and a ResNeXt101 classifier with 16 frames input, is proposed, which has a reaction time of less than three seconds and achieves the highest Levenshtein accuracy (37.5\%) in my experiment. 

\section{Conclusion}
 
In the paper, an online hand gesture recognition framework is presented, which can detect and classify gesture from real-time video stream. The proposed system only has a lag of no more than 3 seconds between performing a gesture and its classification, and can achieve a Levenshtein accuracy of 37.5\% in self-collected dataset. Besides, a systematical evaluation is made from evaluating detector, classifier to evaluating the entire system. Notably, the Levenshtein metric is used, which can evaluate the misclassification, multiple detection and missing detection of the prediction sequence. In addition, several 3D CNN models well trained on Jester database are gained.

All in all, this project is a good attempt to bridge the gap between theory and practical application. Unlike a lot of previous work which just focus on improving the accuracy of the classifier, the project take a holistic consideration on the whole system including reaction time, recognition accuracy and computational cost.

\section{Future Work}
 
The system can still be improved in many aspects. For example, the system is not sensitive to the temporal scale of the gesture, the three-seconds response time of the system cannot satisfy people's expectations, the recognition accuracy of the system is not high enough etc. Despite these flaws, it is possible for us to improve the system by applying more prominent models and finding more refined algorithm.

About the evaluation, it is imperative to find a unified evaluation criteria for online gesture recognition system. Although Levenshtein evaluation metric is used in my experiment, there is no an evaluation metric that is agreed by most researchers so far. In addition, the evaluation in this paper can be further in-depth, such as using larger data sets, adding more comparisons, etc.

Real-world deployment of the system is also a valuable direction. The system is possible to be deployed on vehicles, personal computers and even mobiles, which will be amazing for implementing human computer interface. For example, the system can be extended to mobile phone application to help people with hearing loss, like helping interaction by the sign language, and the system can also be applied to virtual reality gaming to control roles without extra wearable devices.

\section*{Acknowledgment}

The author would like to sincerely thank Dr. Tijana Timotijevic for her supervision and support during this work. This paper was originally written in Autumn 2020 as part of the author's master's dissertation and represents an important first step in the author's research journey.

\bibliographystyle{abbrvnat}
\bibliography{mybib}

\vspace{12pt}

\end{document}